\documentclass[letterpaper]{article} %
\usepackage{aaai2026}  %
\usepackage{times}  %
\usepackage{helvet}  %
\usepackage{courier}  %
\usepackage[hyphens]{url}  %
\usepackage{graphicx} %
\usepackage{natbib}  %
\usepackage{caption} %
\usepackage{algorithm}
\usepackage{algorithmic}
\usepackage{amssymb}
 \usepackage{amsmath} 
 \usepackage{multirow,makecell}
  \usepackage[capitalise,noabbrev]{cleveref}

\usepackage{newfloat}
\usepackage{listings}
\DeclareCaptionStyle{ruled}{labelfont=normalfont,labelsep=colon,strut=off} %
\floatstyle{ruled}
\newfloat{listing}{tb}{lst}{}
\floatname{listing}{Listing}
\nocopyright

\title{Similarity Matters: A Novel Depth-guided Network for Image Restoration and A New Dataset}
\author{
    Junyi He\textsuperscript{\rm 1}, Liuling Chen\textsuperscript{\rm 2}, Hongyang Zhou\textsuperscript{\rm 2},  Xiaoxing Zhang\textsuperscript{\rm 4}, Xiaobin Zhu\textsuperscript{\rm 2}, Shengxiang Yu\textsuperscript{\rm 3},\\ Jingyan Qin\textsuperscript{\rm 1}, Xu-Cheng Yin\textsuperscript{\rm 2}
}
\affiliations{
    \textsuperscript{\rm 1} School of Intelligence Science and Technology, University of Science and Technology Beijing\\
\textsuperscript{\rm 2} School of Computer and Communication Engineering, University of Science and Technology Beijing\\
\textsuperscript{\rm 3} Institute of Botany, the Chinese Academy of Sciences\\
\textsuperscript{\rm 4}	Yizhi, China Telecom\\

   \{d202310391,d202510493\}@xs.ustb.edu.cn, \{ilaopi,qinjingyanking\}@foxmail.com,\{zhuxiaobin,xuchengyin\}@ustb.edu.cn,  zhangxx7@chinatelecom.cn,  yushengxiang@ibcas.ac.cn

}

\usepackage{bibentry}

\begin{document}

\maketitle

\maketitle

\begin{abstract}
Image restoration has seen substantial progress in recent years. 
However, existing methods often neglect depth information, which hurts similarity matching, results in attention distractions in shallow depth-of-field (DoF) scenarios, and excessive enhancement of background content in deep DoF settings. 
To overcome these limitations, we propose a novel \textbf{D}epth-\textbf{G}uided \textbf{N}etwork (DGN)  for image restoration, together with a novel large-scale high-resolution dataset. 
Specifically, the network consists of two interactive branches: a depth estimation branch that provides structural guidance, and an image restoration branch that performs the core restoration task. 
In addition, the image restoration branch exploits intra-object similarity through progressive window-based self-attention and captures inter-object similarity via sparse non-local attention. 
Through joint training, depth features contribute to improved restoration quality, while the enhanced visual features from the restoration branch in turn help refine depth estimation. 
Notably, we also introduce a new dataset for training and evaluation, consisting of \textbf{9,205 high-resolution images} from 403 plant species, with diverse depth and texture variations. 
Extensive experiments show that our method achieves state-of-the-art performance on several standard benchmarks and generalizes well to unseen plant images, demonstrating its effectiveness and robustness.

\end{abstract}

\section{Introduction}

Image restoration aims to recover high-quality images from their degraded counterpart.
Image restoration encompasses sub-tasks such as super-resolution~\cite{zhy1,zhy2}, denoising~\cite{RNINet}, and the removal of weather-induced degradations~\cite{DEMoreNet,DesnowNet}. 
These techniques have been widely adopted in practical applications such as text detection~\cite{zsx1,zsx2,zsx3}, remote sensing~\cite{ZHANG2024107262}, and biomedical imaging~\cite{DBLP:conf/aaai/LyuC0CHHKFLH25}.

\begin{figure}[t]
    \centering
        \resizebox{\linewidth}{!}{
        \includegraphics[width=1\textwidth]{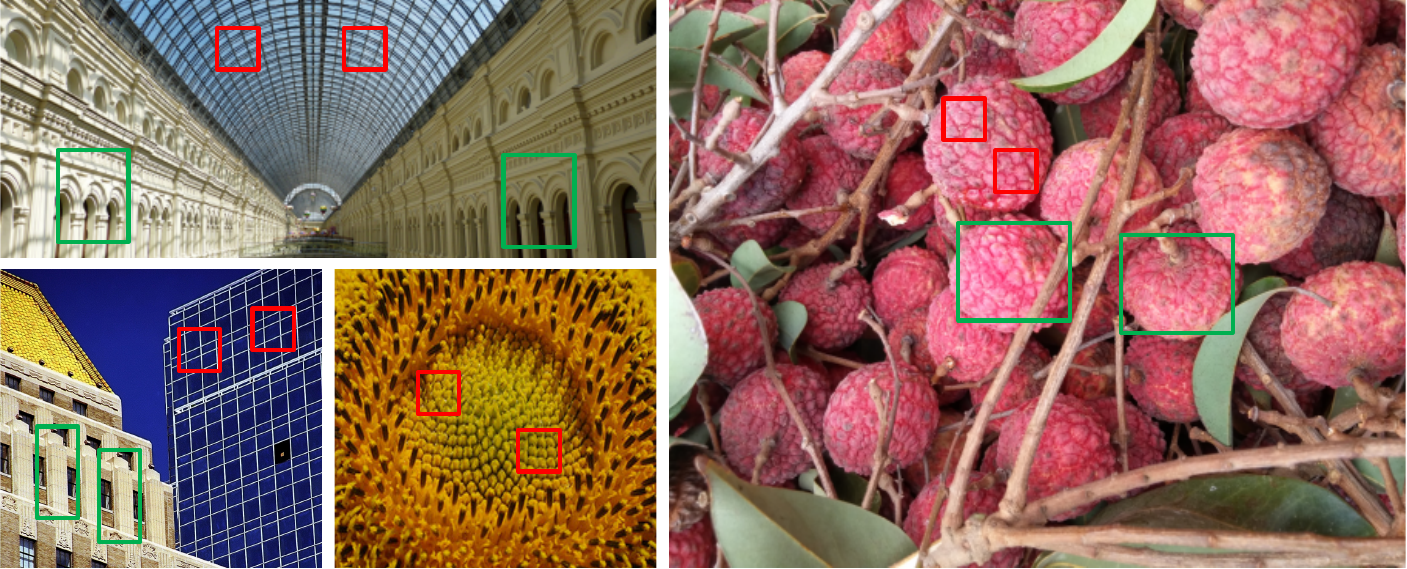}
        }
    \caption{The illustration of intra- and inter-object similarity in natural images, highlighted by red and green rectangles, respectively.} 
    \label{fig_intro}
    \end{figure}

Deep learning has brought remarkable progress to image restoration, with CNN, Transformers~\cite{ViT}, and more recently, Mamba~\cite{Mamba} continuously improving restoration quality and model generalization. 
A key factor enabling this progress is the modeling of visual similarity, which helps preserve fine textures and restore global consistency.
As illustrated in Fig.~\ref{fig_intro}, intra-object similarity refers to the consistency of features within the same semantic region, which helps preserve textures and fine details. Inter-object similarity captures redundancy across different parts of the image and can be leveraged to recover missing or distorted content.
However, most existing methods rely heavily on local convolutional operations or window-based self-attention, limiting their ability to capture long-range dependencies and maintain global coherence.
Meanwhile, depth information has proven particularly effective in weather-related restoration tasks such as image dehazing~\cite{D4}, deraining~\cite{DEMoreNet}, and desnowing~\cite{DesnowNet}, where it serves as a critical prior for distinguishing foreground artifacts from the underlying scene structure.
However, inaccurate or ignored depth perception can introduce unintended effects~\cite{KOALAnet}, including similarity matching errors. These errors can distract attention in shallow depth-of-field (DoF) scenarios by causing salient foreground regions to be overlooked, and also lead to excessive enhancement of background content in deep DoF settings.

\begin{table*}[t]
\centering
 \resizebox{\linewidth}{!}{
\begin{tabular}{|c|c|c|c|c|}
\hline
\textbf{Datasets} & \textbf{Average Resolution} & \textbf{Number of Images} & \textbf{Image Scene} & \textbf{Image Content} \\ \hline
DF2K-P & 1376*2020 & 186 & Indoor & Whole Plant, Stem, Leaf, Flower \\ \hline
PlantSR & 834*1200 & 1030 & Outdoor & Whole Plant, Stem, Leaf, Flower, Fruit, Seed \\ \hline
LSDIR-P & 822*1155 & 3247 & Indoor, Outdoor & Whole Plant, Stem, Leaf, Flower, Fruit, Seed \\ \hline
PlantDR & 1151*1535 & 9205 & Outdoor & Whole Plant, Stem, Leaf, Flower, Fruit, Seed, Root \\ \hline
\end{tabular}
}
\caption{Comparisons of plant image datasets. DF2K-P and LSDIR-P are plants from DIV2K~\cite{DIV2K}/Flickr2K~\cite{MDSR} and LSDIR~\cite{LSDIR}, respectively. PlantSR~\cite{PlantSR} and PlantDR are dedicated plant datasets.}
\label{tab_dinfo}
\end{table*}

To overcome these limitations, we propose a novel \textbf{D}epth-\textbf{G}uided \textbf{N}etwork~(DGN) for image restoration, together with a novel large-scale high-resolution dataset.
One branch focuses on image restoration, while the other estimates depth maps. These two branches are designed to collaborate and guide each other, thereby improving overall restoration performance. The restoration branch captures intra-object similarity through a progressive window-based self-attention mechanism, while inter-object similarity is modeled using a sparse non-local attention module that learns global structural correspondences across repetitive structures.
Notably, we also introduce PlantDR, a new dataset for training and evaluation, featuring 9,205 high-resolution images from 403 plant species. Collected in diverse outdoor settings with diverse depth and texture variations. It spans a comprehensive set of plant parts, including whole plants, flowers, leaves, stems, fruits, seeds and roots.
Our method achieves state-of-the-art~(SOTA) performance in both single image super-resolution~(SR) and denoising tasks, on standard benchmarks and plant-specific datasets. In addition, PlantDR enhances the generalization ability of existing image restoration models on botanical imagery.

Our contributions can be summarized as follows:
\begin{itemize}
  \item We propose a dual-branch restoration network, in which depth estimation guides the image restoration process by providing structural cues. 
  \item We model intra-object and inter-object similarity to enhance the recovery of fine details and global consistency.
  \item We introduce PlantDR, a large-scale, high-resolution dataset for fine-grained plant image restoration, comprising 9205 images with a wide range of structural patterns and depth variations across distinct plant regions.
  \item We verify that our method achieves SOTA performance on standard and plant-specific benchmarks. Moreover, training with PlantDR improves the generalization ability of existing restoration models on botanical images.
\end{itemize}

\begin{figure*}[t]
    \centering
        \includegraphics[width=1\textwidth]{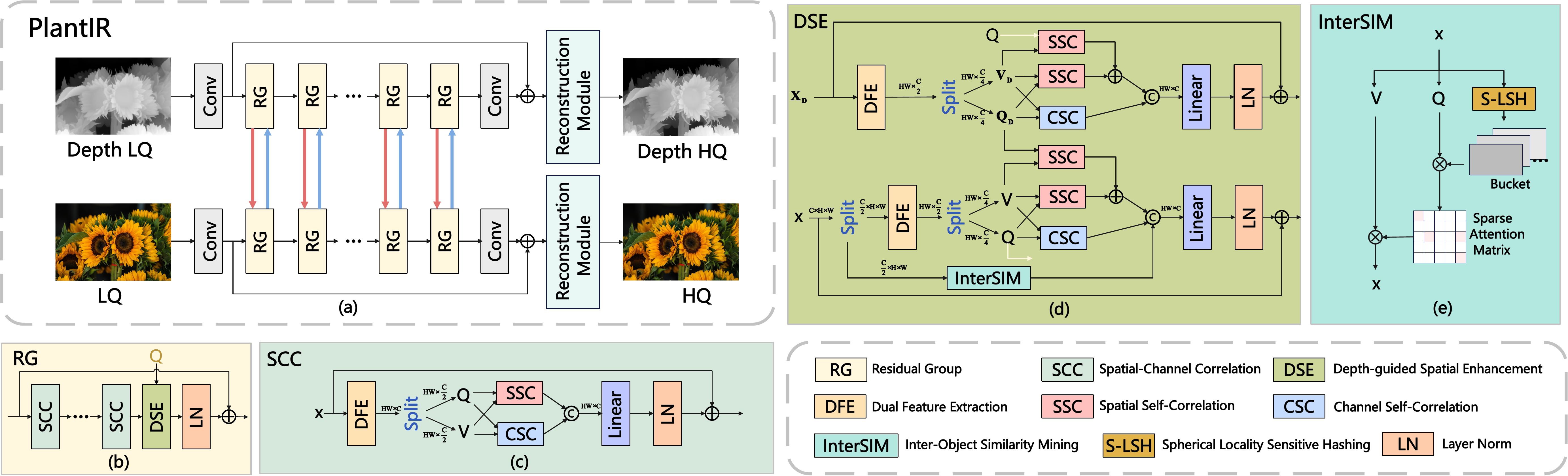} 
    \caption{Illustrations of our method includes: (a) DGN framework, (b) Residual Group structure, (c) the Spatial-Channel Correlation module, (d) Depth-guided Spatial Enhancement module that enables dual-branch information interaction and leverages depth to guide similarity mining, and (e) Inter-Object Similarity Mining module implemented via non-local attention.} \label{fig_framework}
\end{figure*}

\section{Related Work}

\subsection{Existing Image Restoration Methods}

With the rise of deep learning, convolutional neural networks (CNN) have dominated image restoration for several years. Notable CNN-based architectures include DRUNet~\cite{DRUNet}, EDSR~\cite{MDSR}, RCAN~\cite{RCAN} and RNINet~\cite{RNINet}, which have achieved impressive performance by exploiting spatial locality and hierarchical features.
More recently, Transformer-based methods have significantly advanced the field by leveraging long-range dependencies and attention mechanisms. For instance, SwinIR~\cite{SwinIR} integrates the Swin Transformer into an image restoration backbone and demonstrates strong performance across various restoration tasks with a unified architecture. SRFormer~\cite{SRFormer} uses permuted self-attention to achieve high performance with reduced computational burden compared to previous Transformer-based models. Xformer~\cite{Xformer} is a novel hybrid Transformer featuring concurrent spatial-wise and channel-wise Transformer branches coupled by a bidirectional connection unit to enable joint global information modeling across both dimensions. Following this trend, state-space models like Mamba~\cite{Mamba} have emerged as efficient alternatives to self-attention, offering linear-time sequence modeling capabilities. Inspired by Mamba, MambaIR~\cite{MambaIR} adapts Mamba blocks to image restoration, achieving competitive results with improved efficiency. 
These developments reflect a broader shift toward general-purpose, task-agnostic architectures in image restoration. A key factor driving this progress is the effective modeling of visual similarity, which is crucial for preserving fine textures and restoring global consistency.
To achieve this, attention mechanisms play a central role in identifying and exploiting visual similarity. Spatial attention~\cite{SAN}, channel attention~\cite{RCAN}, self-attention~\cite{SwinIR}, and non-local attention~\cite{NLSN} are designed to enhance feature representation and relational modeling. Among them, self-attention and non-local modules compute pairwise feature similarities, followed by softmax normalization to emphasize relevant contextual information across the image.
Due to computational constraints, they are often applied locally, limiting global-local feature integration. 
NLSN~\cite{NLSN} reduces complexity using spherical locality sensitive hashing, while HiTSR~\cite{HiTSR} adopts correlation maps instead of attention maps. To mine both intra-object and inter-object similarities, we integrate NLSN and HiTSR in parallel and use depth maps to handle complex depth variations.

\subsection{Image Restoration Datasets}

In image restoration tasks, datasets are essential for driving model performance. 
 High-resolution, diverse, and large-scale datasets can significantly enhance model performance without modifying the underlying architecture~\cite{LSDIR}. Currently, datasets such as DIV2K~\cite{DIV2K} and Flickr2K~\cite{MDSR} are widely adopted due to their 2K resolution. However, as models increase in complexity and restoration tasks become more demanding, the limited resolution and the total of only 3,450 images are no longer sufficient. To address this limitation, DIV8K~\cite{DIV8K} introduced higher-resolution images, and LSDIR~\cite{LSDIR} further expanded the dataset scale by providing 84,991 images of natural scenes.
Nevertheless, existing datasets fail to capture the depth variations commonly present in real-world images. Recent research has even attempted to remove the intentional blur found in some datasets~\cite{HGGT}, and this assumption exacerbates the model’s insensitivity to depth information.
Plant-related images typically exhibit both rich textures and diverse depth characteristics. However, as listed in Table~\ref{tab_dinfo}, current image restoration datasets contain only a limited number of plant images. Specifically, the DF2K dataset includes only 186 plant images out of 3,450 (800 from DIV2K and 2,650 from Flickr2K), while the LSDIR dataset contains merely 3,247 plant images among 84,992. Even PlantSR, a dataset specifically curated for plant image super-resolution, includes only 1,080 images.
To overcome these limitations, we propose a large-scale plant dataset that offers high-resolution images with rich textures and diverse depth characteristics. The proposed dataset is intended to enhance image restoration models by enabling them to learn depth-aware features directly from the data.

\section{Proposed Method}

To address the challenge of varying depth-of-field in real-world image restoration, we propose a dual-branch network where a depth estimation branch provides structural guidance to the restoration branch. We first introduce the overall framework, then detail the similarity mining modules, followed by the depth-guided spatial enhancement module and the training objectives.

\textbf{Framework.} 
The architecture of our proposed DGN is illustrated in Fig.~\ref{fig_framework} (a). The model consists of two parallel branches with mirrored structures, which exchange features during the processing of each residual group (RG) to facilitate mutual guidance and feature enhancement.
Without loss of generality, we take the image restoration branch as an example for illustration. 
This branch can be divided into three stages: the shallow feature extraction module, the deep feature extraction module, and the reconstruction module. The shallow feature extraction module uses a single convolutional layer to project the input image into the feature space.
The deep feature extraction module consists of $N$ RGs, followed by a convolutional layer and a long-term residual connection. Each RG receives complementary features from the parallel branch to enhance its own feature learning.
The reconstruction module employs upsampling layers for SR tasks, while a standard convolutional layer is used for other image restoration tasks.
Let $x, x_D \in \mathbb{R}^{H/s \times W/s \times 3}$ denote the input low-quality~(LQ) image and low-quality depth~(depth LQ) image, respectively, where $s$ is the scale factor. In SR tasks, $s \neq 1$, while in other tasks, $s = 1$. The output images are denoted as $y, y_D \in \mathbb{R}^{H \times W \times 3}$. The overall framework can be formulated as:
\begin{align}
f,f_D &=Conv(x),Conv(x_D), \\
f',f_D' &= RG_N(RG_{N-1}(..RG_1(f,f_D))), \\
y,y_D &= RD(f'),  RD(f_D'), 
\end{align}
where $RD(\cdot)$ is the reconstruction module.

\textbf{Intra-Object and Inter-Object Similarity Mining Module.} 
We adopt HiTSR~\cite{HiTSR} as the backbone of our framework. HiTSR introduces a Spatial-Channel Correlation (SCC) module, illustrated in Fig.~\ref{fig_framework} (b). The SCC module consists of three components: Dual Feature Extraction (DFE), Spatial Self-Correlation (SSC), and Channel Self-Correlation (CSC).
DFE uses convolutional and linear layers within a gating mechanism to extract crucial spatial and channel features, formulated as:
\begin{equation}
x_{\text{dfe}} = Conv(x) \odot L(x),
\end{equation}
where $x_{\text{dfe}}$ is the output of the DFE module, $Conv(\cdot)$ denotes a bottleneck convolutional block, and $L$ is a linear projection layer.
In traditional self-attention, the input $x$ is projected into query, key, and value via linear transformations, followed by attention computation using $QK^{T}$ and aggregation via multiplication with $V$. In contrast, both SSC and CSC modules employ correlation maps instead of attention maps for information aggregation. To reduce computational overhead, the input $x$ is split directly along the channel dimension into query and value tensors. Correlation maps are computed based on these splits, and the aggregated result is obtained via multiplication with $V$.
The computation is expressed as:
\begin{align}
\text{SSC}(Q_i, V_i) &= \left( \frac{Q_i V_i^\top}{\sqrt{C/2}} + B \right) \cdot V_i, \\
\text{CSC}(Q_i, V_i) &= \left( \frac{Q_i V_i^\top}{\sqrt{h_i \times w_i}} \right) \cdot V_i, 
\end{align}
where $Q_i, V_i \in \mathbb{R}^{h_i w_i \times \frac{C}{2}}$ denote the $i$-th non-overlapping windows split from $Q$ and $V$, and $B$ represents relative positional encoding. HiTSR gradually increases the window size, effectively expanding the receptive field. SSC operates in the spatial domain, while CSC captures correlations along the channel dimension. For more detail, please refer to HiTSR for more architectural details.

In our proposed DGN framework, we define the combination of SSC and CSC as the Intra-object Similarity Mining (IntraSIM) module, since their limited receptive fields within local windows are well-suited for capturing fine-grained local similarities.
To complement this, we incorporate a Non-local Sparse Attention mechanism~\cite{NLSN} as the Inter-object Similarity Mining (InterSIM) module, which captures global similarities across the entire image. Traditional non-local attention requires computing pairwise similarities over the full spatial resolution, resulting in quadratic complexity that is prohibitive for high-resolution inputs.
To reduce this cost, InterSIM uses Spherical Locality Sensitive Hashing (S-LSH) to cluster similar features into buckets, computing similarities only within each bucket. This enables efficient modeling of long-range dependencies with significantly reduced computational overhead.

\subsection{Depth-guided Spatial Enhancement Module}

The Depth-Guided Spatial Enhancement (DSE) module, illustrated in Fig.~\ref{fig_framework}(d), demonstrates how the InterSIM and IntraSIM components operate cooperatively and how the dual-branch pathways interact to enhance feature learning.
In the image restoration branch, the input feature map is first split evenly along the channel dimension. One half is fed into the IntraSIM module, and the other into the InterSIM module.
The depth feature map is designed to have half the number of channels as the image feature, matching the dimensionality of each split. These three components are denoted as $X_1$, $X_2$, and $X_D$, respectively.
Both $X_2$ and $X_D$ are first refined using the DFE mechanism and then split into query and value components for SCC computation.
Cross-branch interaction is introduced by injecting the depth query $Q_D$ into the image branch and the image query $Q$ into the depth branch. 
These cross-branch queries are paired with the corresponding values ($V$ and $V_D$) to compute spatial self-correlation. The resulting SSC features are combined with intra-branch SSC results, concatenated with the CSC outputs, and passed through a linear layer to capture intra-object similarity.
In the image-restoration branch, the InterSIM output is concatenated before the linear projection, fusing both intra- and inter-object similarity information. CSC interaction is omitted across branches because depth provides only spatial guidance rather than channel-wise information to the image features.
The full forward process of the DSE module is detailed in Algorithm~\ref{alg_dse}, where $\mathrm{Split}(\cdot)$ denotes channel-wise feature splitting and $[\cdot,\cdot]$ represents feature concatenation. To mitigate the distribution gap between depth and image features, a normalization layer is applied after the DSE module within each residual group to stabilize training.

\begin{algorithm}[t]
\caption{DSE module Forward Pass}
\label{alg_dse}
\textbf{Input}: Image feature $X$, Depth image feature $X_D$\\
\textbf{Output}: New image feature $X$, new depth image feature $X_D$
\begin{algorithmic}[1]
\STATE Split $X$ into $X_1$, $X_2$
\STATE $X_D', X_2' \leftarrow \mathrm{DFE}(X_D), \mathrm{DFE}(X_2)$
\STATE Split $X_D'$ into $Q_D$, $V_D$
\STATE Split $X_2'$ into $Q$, $V$
\STATE $X_1' \leftarrow \mathrm{InterSIM}(X_1)$
\STATE $T \leftarrow [\mathrm{CSC}(Q, V), \mathrm{SSC}(Q, V) + \mathrm{SSC}(Q_D, V), X_1']$
\STATE $T_D \leftarrow [\mathrm{CSC}(Q_D, V_D), \mathrm{SSC}(Q_D, V_D) + \mathrm{SSC}(Q, V_D)]$
\STATE $X \leftarrow \mathrm{Linear}(\mathrm{LN}(T)) + X$
\STATE $X_D \leftarrow \mathrm{Linear}(\mathrm{LN}(T_D)) + X_D$
\end{algorithmic}
\end{algorithm}

\subsection{Training Objective}

We use an image loss to ensure global consistency. For depth maps, an $l_1$ loss enforces overall consistency, complemented by an affine-invariant depth loss~\cite{DepthAnything} that removes scale and translation ambiguities, focusing on relative depth.

\noindent\textbf{Image loss.} The image loss between reconstructed image $y$ and HQ image $x$ can be formulated as:
\begin{equation}
\mathcal{L}_{image}=l_1(y,x),
\end{equation}
where $l_1(\cdot)$ denotes the popular L1 loss.

\noindent\textbf{Depth loss.} Depth loss between reconstructed depth image $y_D$ and HQ depth image $x_d$ can be formulated as:
\begin{equation}
\mathcal{L} _{depth}=\lambda_1\mathcal{L}_{1}(y_D,x_D)+\lambda_2\mathcal{L}_{AID}(y_D,y),
\end{equation}
where $\mathcal{L}_{AID}(\cdot)$ denotes affine-invariant depth loss~\cite{DepthAnything}.

The total loss $\mathcal{L} _{total}=\mathcal{L} _{image}+\mathcal{L} _{depth}$, where the weights for each loss are set as: $\lambda_1=\lambda_2=0.01$.

\begin{figure}[t]
    \centering
        \resizebox{\linewidth}{!}{
        \includegraphics[width=1\textwidth]{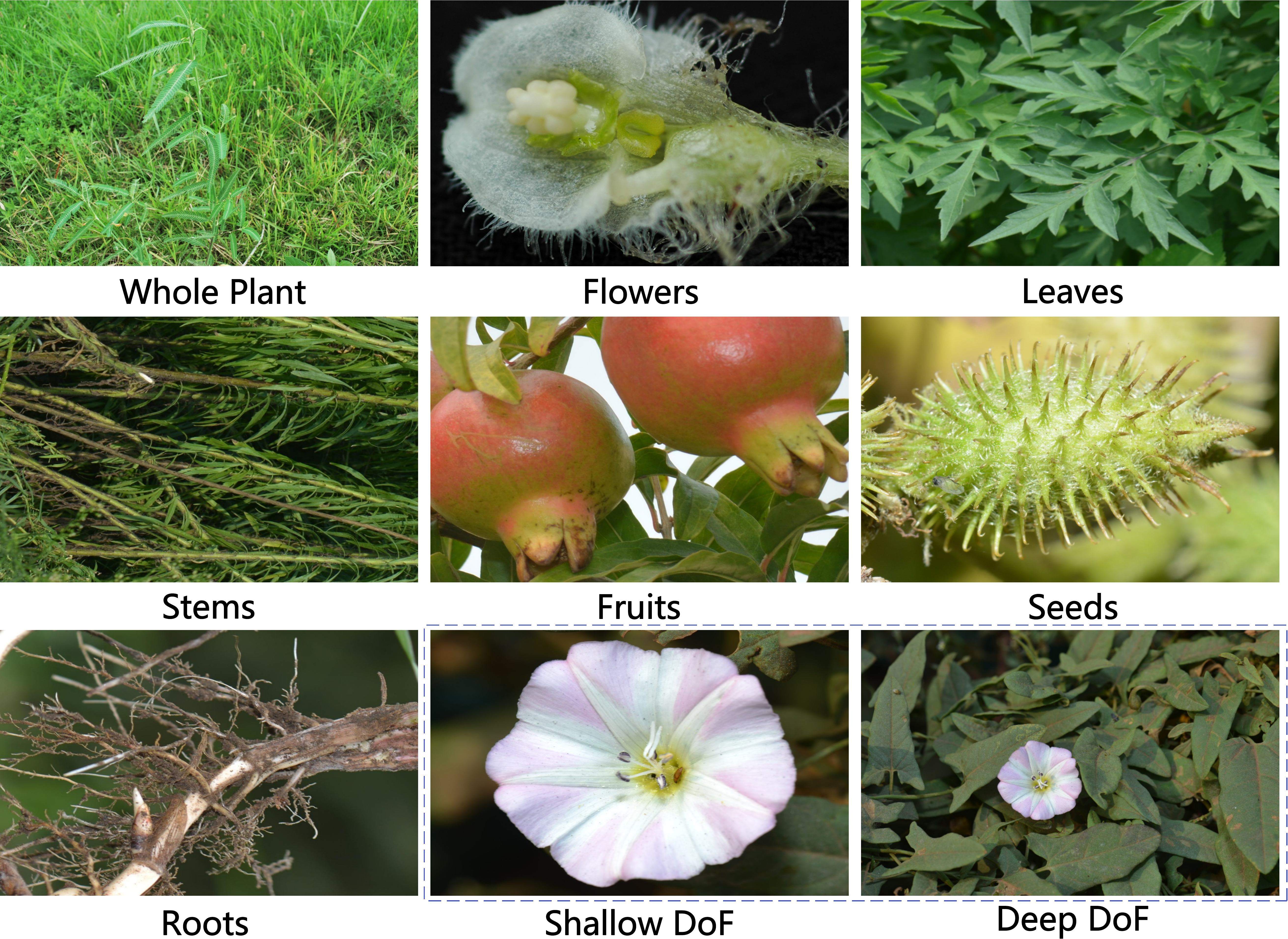}
        }
    \caption{Composition and characteristics of the Plant Image Restoration Dataset (PlantDR). Blue dashed boxes highlight shallow and deep depth-of-field examples.} 
    \label{fig_orisize}
    \end{figure}

\section{Plant Dataset for Image Restoration}

To fully demonstrate the effectiveness of our proposed DGN, we introduce PlantDR, a high-resolution plant-image dataset tailored for image restoration research. PlantDR contains images with diverse depths of field and rich fine-grained textures and structures typical of various plant species. It captures abundant intra- and inter-object similarities and covers a wide range of plant types with complex surface patterns.

\subsection{Benchmark Collection}

\textbf{Image Acquisition.} We collected a total of 15,999 high-resolution images of 441 invasive plant species in China using professional digital cameras such as the Nikon D800E and Nikon D90. The average image resolution is $4509 \times 6746$ pixels, providing rich visual details suitable for image restoration tasks. As illustrated in Fig.~\ref{fig_orisize}, the dataset covers a wide range of plant parts, including whole plants, flowers, leaves, stems, fruits, seeds and roots, capturing diverse and fine-grained textures.
To ensure visual clarity and focus on structural information, most images were taken with a shallow depth of field, emphasizing the surface patterns and morphological features of individual plant components. The dataset was collected under natural lighting conditions across different seasons and regions, contributing to its diversity in color, illumination, and species appearance.

\textbf{Duplicate Removal.} To reduce data redundancy, we implemented a perceptual hashing-based duplicate removal algorithm~\cite{zauner2010phash}. 
This method preserves visual similarity while being robust to minor transformations such as brightness changes or compression.
For each image, a perceptual hash value was computed through the following steps: grayscale conversion, optional brightness normalization, resizing to $8 \times 8$, and thresholding by average pixel intensity. The resulting binary matrix was flattened into a 64-bit hash $h \in \{0,1\}^{64}$ and converted to an integer representation.

The similarity between images was measured using the Hamming distance $d_H(h_1, h_2)$, defined as:
\begin{equation}
d_H(h_1, h_2) = \sum_{i=1}^{64} \mathbf{1}[h_1^{(i)} \ne h_2^{(i)}],
\end{equation}
where $h_1$ and $h_2$ denote the perceptual hash values of two images, $\mathbf{1}[\cdot]$ is the indicator function that returns 1 if the condition inside is true, and 0 otherwise. 
Images with $d_H(h_1, h_2) < \delta$ (we used $\delta = 10$) were considered visually similar. For each category, only one image per near-duplicate set was kept to preserve intra-class diversity.

Furthermore, low-light images were excluded based on average brightness. An image was considered too dark and discarded if its mean grayscale intensity was below a fixed threshold (e.g., 40). This ensured that the dataset retained only high-clarity and well-lit images suitable for further processing. After applying the duplicate removal and low-light filtering algorithms, the dataset was reduced from 15,999 to 8,104 images. The average resolution of the retained images is approximately $4338 \times 6483$ pixels, preserving rich texture information essential for fine-grained restoration tasks.

\textbf{Manual Filtering and Patch Generation.}
To further eliminate near-duplicate images and low-light samples that may have been missed by the automated filtering, three annotators manually inspected all remaining images. 
After this manual refinement process, the dataset was reduced to 3,261 high-resolution images covering 388 distinct plant categories, with an average resolution of $4131 \times 6159$ pixels.
The images were then partitioned using two distinct methods. First, they were systematically divided into non-overlapping patches of size $1535 \times 1151$, producing 63,389 patches. Second, ten annotators manually cropped patches from the 3,261 images, resulting in 7,432 patches.
Patches derived from both methodologies may contain suboptimal regions, including blurred backgrounds, damaged borders, or other artifacts. Furthermore, potential duplication exists between patches generated by the two partitioning approaches. Consequently, subsequent processing stages require additional cleaning to address these issues.

\begin{figure*}[t]
    \centering
        \resizebox{\linewidth}{!}{
        \includegraphics[width=1\textwidth]{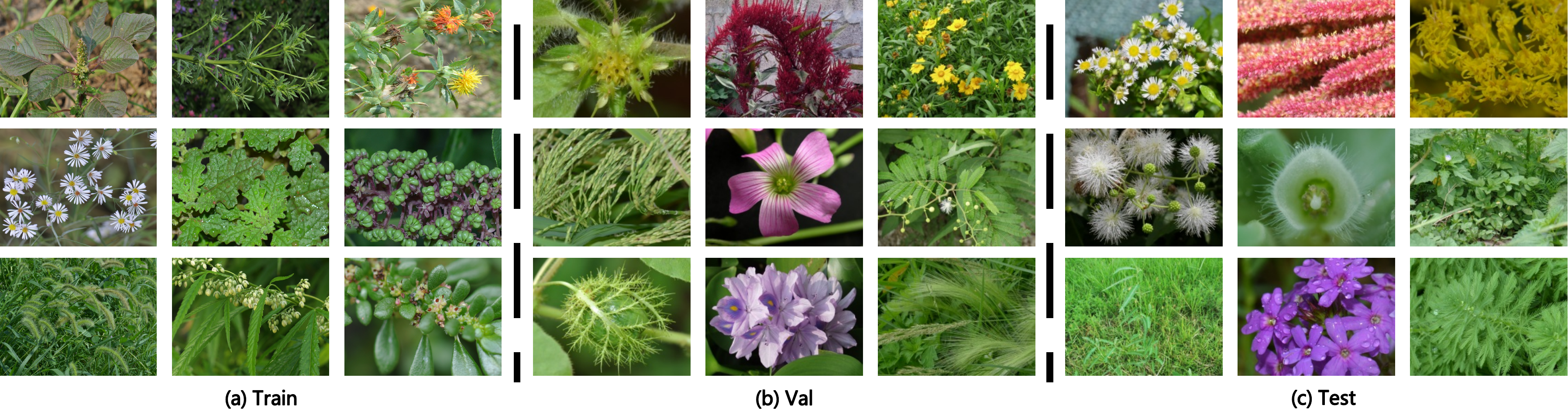}
        }
    \caption{Sample images from the training, validation, and test sets in the PlantDR dataset.} 
    \label{fig_plantDR_split}
    \end{figure*}

\textbf{High-Quality Image Selection.}
To ensure the final dataset contains only visually clear and texture-rich images, we implemented a rigorous two-phase annotation process. 
Ten annotators first independently evaluated the filtered data within each category. This was followed by a cross-validation phase where disagreements were resolved to ensure consistency. Images exhibiting visible blur, out-of-focus regions, or insufficient texture detail were manually excluded during this process.
Following this quality control protocol, a total of 9,205 high-resolution images ($1535 \times 1151$) were retained. These images exhibit well-defined structures and clear textures, rendering them suitable for fine-grained image restoration tasks.
Finally, the dataset was partitioned into non-overlapping sets: 9,005 images for training, 100 for validation, and 100 for testing. Strict measures were implemented to prevent image overlap between the training set and the validation/test sets.
Fig.~\ref{fig_plantDR_split} shows representative samples from our PlantDR dataset, especially in the test set, which includes images with both shallow depth-of-field and deep depth-of-field.

\begin{figure*}[t]
    \centering
        \resizebox{\linewidth}{!}{
        \includegraphics[width=1\textwidth]{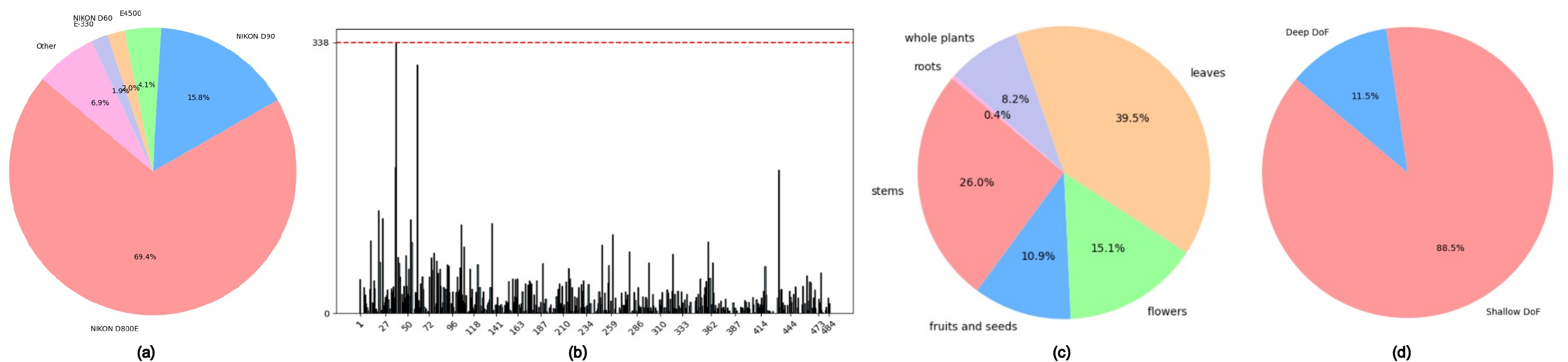}
        }
    \caption{Overview of the dataset characteristics: (a) Distribution of camera types used for image acquisition. (b) Diversity of plant species included in the dataset~(388 species; IDs without images excluded). (c) Variety of plant parts captured, such as whole plants, flowers, leaves, stems, fruits, seeds and roots. (d) Proportions of shallow and deep depth-of-field images.} 
    \label{fig_plotchart}
    \end{figure*}

\subsection{Benchmark characteristics}

Our dataset exhibits three notable characteristics:

\textbf{Captured with a diverse range of cameras.} As shown in Fig.~\ref{fig_plotchart}(a), the images were captured using 24 different camera models under natural scenes, introducing real-world domain shifts. The NIKON D800E and D90 account for 69.4\% and 15.8\% of the images respectively, with the remaining 14.8\% captured by 22 other camera models. This diversity provides a strong foundation for evaluating generalization and domain adaptation.

\textbf{Rich diversity in plant species and plant parts.} As illustrated in Fig.~\ref{fig_plotchart}(b-c), the dataset includes 388 unique plant species.
It also includes diverse plant parts such as flowers, leaves, stems, fruits, seeds and roots, supporting fine-grained texture modeling and restoration.

\textbf{Extensive variation in depth of field.} As shown in Fig.~\ref{fig_plotchart}(d), the dataset includes images with diverse depth-of-field, including 88.5\% shallow depth-of-field images and 11.5\% deep depth-of-field images. This variation allows models trained on our dataset to adaptively learn and leverage depth-related information.

\begin{figure*}[t]
    \centering
        \resizebox{\linewidth}{!}{
        \includegraphics[width=1\textwidth]{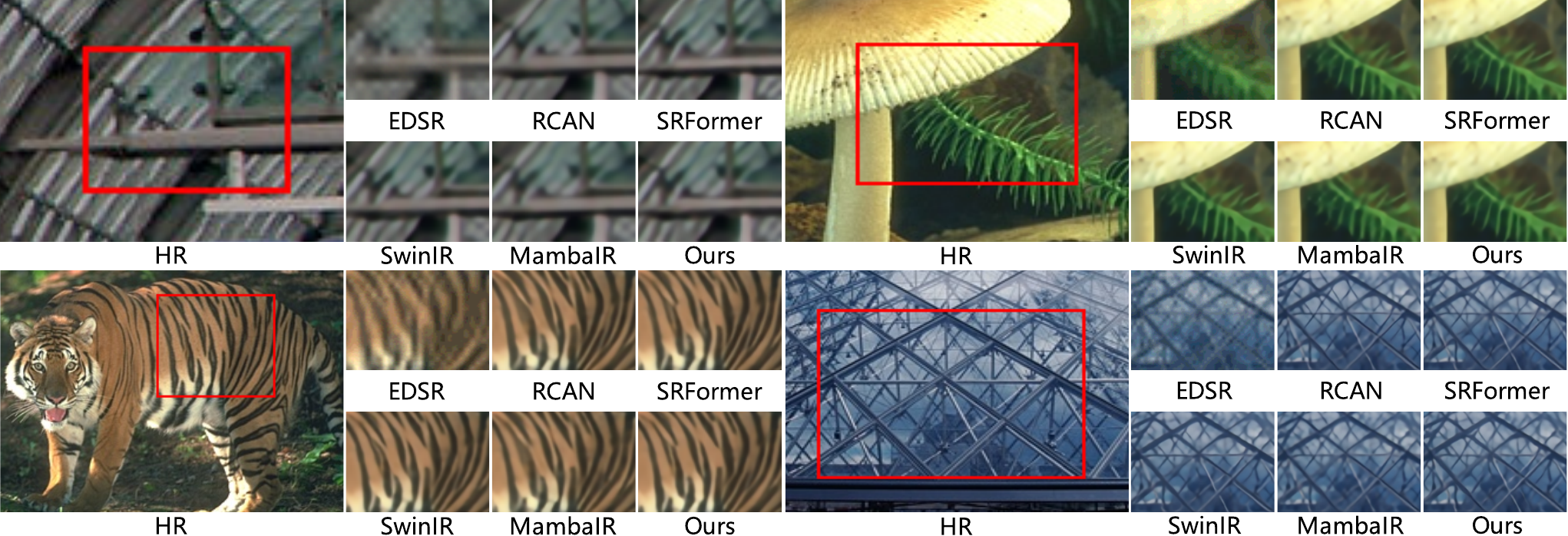}
        }
    \caption{Visual comparisons on common benchmarks for the SR task.} 
    \label{fig_bicubic}
    \end{figure*}

\renewcommand{\arraystretch}{1.2} %
\begin{table}[!ht]

\centering

\small
\setlength{\tabcolsep}{2mm}
\resizebox{\linewidth}{!}{

\begin{tabular} {c|c|c|cccccc}
    \hline
    Methods   & Scale &  Metrics &  Set5 & Set14 & B100 & Urban100 & Mamga109 & DIV2K-VAL \\  
\hline
\multirowcell{3}{\textbf{EDSR}} & $\times 4$  &PSNR$\uparrow$ &27.61 &25.72& 23.26&25.74 &24.98&27.96	 \\
& $\times 4$   &SSIM$\uparrow$  &0.7851 &0.6975 &0.6657 & 0.6459	&0.7851 &7744\\
1.52M & $\times 4$     &LPIPS$\downarrow$  & 0.2714& 0.3956  & 0.4339 & 0.4377 & 0.3066	&0.3831 \\

\hline
\multirowcell{3}{\textbf{RCAN}}& $\times 4$    &PSNR$\uparrow$  &30.92&27.91 &27.39 &25.36 &28.97 	& 30.07 \\
& $\times 4$ &SSIM$\uparrow$  &0.8782 &0.7710  &0.7266 &0.7638	&0.8863 &0.8269\\
15.59M& $\times 4$ &LPIPS$\downarrow$ &0.2012 &0.3096 &0.3975  &0.2778 &0.1536 	& 0.2986\\

\hline
\multirowcell{3}{\textbf{SRFormer}}& $\times 4$    &PSNR$\uparrow$ &31.14  &28.02 &\textbf{27.53} &\textbf{25.77} &\textbf{29.40} 	&\textbf{30.28} \\
& $\times 4$ &SSIM$\uparrow$ & 0.8818& 0.7747&\textbf{ 0.7314}& \textbf{0.7804}  & \textbf{0.8949}	&0.8333 \\
10.54M& $\times 4$ &LPIPS$\downarrow$  &0.1963 &0.3021 &0.3891 &0.2546 &0.1424 	&0.2852 \\
 
\hline
\multirowcell{3}{\textbf{SwinIR}}& $\times 4$    &PSNR$\uparrow$ &\textbf{31.24} &\textbf{28.04}  &27.50 &25.68 &29.16 	&30.26 \\
& $\times 4$ &SSIM$\uparrow$  & \textbf{0.8835} & \textbf{0.7754} &0.7313&0.7781 &0.8914 	& 0.8332\\
13.29M& $\times 4$ &LPIPS$\downarrow$ &\textbf{0.1925} &0.3027  &0.3894&0.2585 &0.1489 & 0.2858\\

\hline
\multirowcell{3}{\textbf{MambaIR}}& $\times 4$    &PSNR$\uparrow$  &30.97  &27.97 &27.42 &25.50 & 29.29	&30.09 \\
& $\times 4$ &SSIM$\uparrow$ &0.8799 &0.7731 &0.7284 &0.7699 &0.8914 	&0.8276 \\
20.57M& $\times 4$ &LPIPS$\downarrow$ &0.1937 &\textbf{0.2956} &\textbf{0.3851} &0.2598 &\textbf{0.1394} 	&0.2926 \\

\hline
\multirowcell{3}{\textbf{Ours}}& $\times 4$    &PSNR$\uparrow$  &31.06 &27.98 &27.46 &25.73 &29.25 	&\textbf{30.28} \\
& $\times 4$ &SSIM$\uparrow$ &0.8816 &0.7736 &0.7293 &0.7786 &0.8934 	&\textbf{0.8335 }\\
3.52M& $\times 4$ &LPIPS$\downarrow$ &0.1982 & 0.3038& 0.3904& \textbf{0.2495} &0.1411 	&\textbf{0.2787} \\

\hline

\end{tabular}{}
}
    \caption{ Quantitative comparison with state-of-the-art methods on common benchmarks.}
\label{tab_sr_com}
\end{table}

\renewcommand{\arraystretch}{1.2} %
\begin{table}[!ht]

\centering

\small
\setlength{\tabcolsep}{2mm}
\resizebox{\linewidth}{!}{

\begin{tabular} {c|c|c|ccc|ccc}
    \hline
    Train $\rightarrow$ Test   & Scale & Metrics  &  P1 $\rightarrow$ P1 &  P2 $\rightarrow$ P1  & P3 $\rightarrow$ P1  &   P1 $\rightarrow$ P3 & P2 $\rightarrow$ P3 & P3 $\rightarrow$ P3\\  
\hline
\multirowcell{3}{\textbf{EDSR}} & $\times 4$  &PSNR$\uparrow$  &31.93  &33.82 & \textbf{34.05}&31.94&34.30&\textbf{34.56} \\ 
& $\times 4$   &SSIM$\uparrow$  &0.8869&0.8958 &\textbf{0.8988}  &0.8922 	&0.8993 &\textbf{0.9018}\\ 
1.52M& $\times 4$     &LPIPS$\downarrow$  &0.2491  &0.2217 &\textbf{0.2109}  &0.2485 	&0.2243  &\textbf{0.2091}\\

\hline
\multirowcell{3}{\textbf{RCAN}}& $\times 4$    &PSNR$\uparrow$  &36.22 &36.29 &\textbf{36.30}   &36.93	&37.03&\textbf{37.15} \\ 
& $\times 4$ &SSIM$\uparrow$  &0.9219  &\textbf{0.9224} &0.9221  	&0.9298 &0.9314&\textbf{0.9330}\\
15.59M& $\times 4$ &LPIPS$\downarrow$  &0.1692 &0.1624 &\textbf{0.1610} 	&0.1648&0.1587& \textbf{0.1504}\\

\hline
\multirowcell{3}{\textbf{SRFormer}}& $\times 4$    &PSNR$\uparrow$  &35.95  &36.32 &\textbf{36.35}  &36.82 &37.16 &\textbf{37.32}	 \\
& $\times 4$ &SSIM$\uparrow$ &0.9219 &\textbf{0.9235} &0.9234  &0.9310 &0.9315 &\textbf{0.9346} \\
10.54M& $\times 4$ &LPIPS$\downarrow$  &0.1655 &\textbf{0.1515} &0.1622 	&0.1558 &\textbf{0.1423} &0.1460  \\
 
\hline
\multirowcell{3}{\textbf{SwinIR}}& $\times 4$    &PSNR$\uparrow$  &36.15 &36.12 &\textbf{36.21}	&36.90&36.91&\textbf{37.16} \\
& $\times 4$ &SSIM$\uparrow$    &0.9215 &0.9221 &\textbf{0.9224}  &0.9296&0.9313& \textbf{0.9331}\\
13.29M& $\times 4$ &LPIPS$\downarrow$  &0.1618 &\textbf{0.1563} &0.1644 	&0.1525&\textbf{0.1465}&0.1480\\

\hline
\multirowcell{3}{\textbf{MambaIR}}& $\times 4$    &PSNR$\uparrow$   &36.18 &36.20 &\textbf{36.23}&37.10 &37.12&\textbf{37.15} \\
& $\times 4$ &SSIM$\uparrow$ &0.9210 &0.9212 &\textbf{0.9217}&0.9289&0.9312& \textbf{0.9314}\\
20.57M& $\times 4$ &LPIPS$\downarrow$ &0.1727 &\textbf{0.1633} & 0.1654	&0.1639&\textbf{0.1580}&0.1679 \\

\hline
\multirowcell{3}{\textbf{Ours}}& $\times 4$    &PSNR$\uparrow$  &36.28 & 36.30&\textbf{36.33}  &37.22&37.22	&\textbf{37.30} \\
& $\times 4$ &SSIM$\uparrow$  &0.9223 &0.9225 &\textbf{0.9230} 	&0.9327&0.9333&\textbf{0.9335} \\
3.52M& $\times 4$ &LPIPS$\downarrow$ &0.1602 &\textbf{0.1574} &0.1596 &0.1505&\textbf{0.1452}&0.1487 \\

\hline

\end{tabular}{}
}
    \caption{ Quantitative comparison with state-of-the-art methods on plant benchmarks. P1 denotes PlantSR~\cite{PlantSR}, P2 denotes LSDIR-plant~\cite{LSDIR}, and P3 refers to our proposed PlantDR dataset.}
\label{tab_sr_plant}
\end{table}

\setlength{\tabcolsep}{1.2pt} %
\renewcommand{\arraystretch}{1} %
\begin{table}[t]
\centering

\small

      \centering
      \small
      \setlength{\tabcolsep}{2mm}
      \begin{tabular}{c|c|c|c}
        \hline
        Setting & PSNR$\uparrow$ & SSIM$\uparrow$ & LPIPS$\downarrow$ \\
        \hline
         w/o Depth & 30.24 & 0.8324 & 0.2800 \\
        \hline
        w Depth & \textbf{30.28} & \textbf{0.8335} & \textbf{0.2787} \\
        \hline
      \end{tabular}
\caption{Ablation study of dual branch settings.}
      \label{tab_abDual}
\end{table}

\section{Experiments}

\subsection{Experimental Settings}

We conduct experiments on two low-level vision tasks: single-image super-resolution (scale factor equals 4) and image denoising (See in \textbf{Suppl. Mat.}).
We employ DIV2K~\cite{DIV2K} and Flickr2K~\cite{MDSR} to pretrain our DGN model and then trained with plant dataset, PlantSR~\cite{PlantSR}, LSDIR-plant~\cite{LSDIR} and our PlantDR. We evaluate the model's performance on both common benchmarks and plant datasets.
For common benchmark, we test on Set5~\cite{Set5}, Set14~\cite{Set14}, Urban100~\cite{Urban100}, B100~\cite{B100}, Manga109~\cite{Manga109}, and DIV2K validation set~\cite{DIV2K}.
For plant datasets, we tested on PlantDR testset and PlantSR~\cite{PlantSR} testset.
We adopt Peak Signal-to-Noise Ratio~(PSNR), Structural Similarity Index (SSIM)~\cite{SSIM}, Learned Perceptual Image Patch Similarity (LPIPS)~\cite{LPIPS} to quality the model performance.
We utilize a single NVIDIA RTX A6000 for all training and inference. We randomly sample 8 RGB image patches of $256 \times 256$ as the inputs during training. These patches are augmented with random rotations of 90$^{\circ}$, 180$^{\circ}$, 270$^{\circ}$ and horizontal flips. We choose ADAM optimizer \cite{ADAM} with $\beta_1=0.9$, $\beta_2=0.999$ to train the model for 500K iterations. The initial learning rate is set as $3 \times 10^{-4}$ and subsequently halved at the {250k,400k,450k,475k}-th iterations.
We set $N=6$, $M=6$. An $8 \times 8$ base window is used for SCC, with hierarchical ratios [0.5,1,2,4,6,8] defined relative to this base size within each RG.
The depth low quality image is generated by Depth Anything V2~\cite{DepthAnythingV2}.

\subsection{Performance of Super-Resoltuion}
The five State-of-the-Art methods compared are EDSR~\cite{MDSR}, RCAN~\cite{RCAN}, SRFormer~\cite{SRFormer}, SwinIR~\cite{SwinIR} and MambaIR~\cite{MambaIR}.
Among them, EDSR, RACN, and SRFormer are representative SOTA methods in the SR task employing CNN and Transformer architectures, while SwinIR and MambaIR are Image Restoration models utilizing Transformer and Mamba architectures, respectively.
We trained all the models from scratch using DF2K to ensure a fair comparison. All other hyper-parameters were kept identical to their original configurations.
We first test on a common benchmark, and then compare in plant datasets.

\textbf{Comparision on Common Benchmarks.}
The quantitative and qualitative results are presented in Tab.~\ref{tab_sr_com} and Fig.~\ref{fig_bicubic}, respectively.
Quantitative evaluations show that our method achieves SOTA perceptual quality with only one-second the model size of SRFormer. This is evidenced by the best LPIPS scores on Urban100 (0.2495) and DIV2K-VAL (0.2787), while still maintaining competitive PSNR and SSIM performance. 
Our model performs particularly well on Urban100, where the abundance of intra- and inter-object similarities aligns well with our design objective focused on depth-guided similarity mining.
Qualitative evaluations show that our model reconstructs plausible textures even in blurred regions, mainly due to the depth-guided auxiliary estimation module.

\textbf{Comparision on Plant Benchmarks.}
Tab.~\ref{tab_sr_plant} establishes a cross-dataset evaluation framework using three Plant benchmarks: PlantSR (P1), LSDIR-plant (P2), and our PlantDR (P3). This design assesses  in-domain test such as P3 to P3, and cross-domain generalization test including P3 to P1. 
Quantitative evaluations show that our method achieves SOTA perceptual quality with only one-second the model size of SRFormer. 
Moreover, results show that our PlantDR dataset benefits not only our model but also improves baseline performance in both in-domain and cross-domain settings. Models trained on PlantDR achieve consistently better LPIPS and PSNR scores, indicating that its diverse, well-structured content enhances generalization and robustness.

\begin{figure*}[t]
    \centering
        \resizebox{\linewidth}{!}{
        \includegraphics[width=1\textwidth]{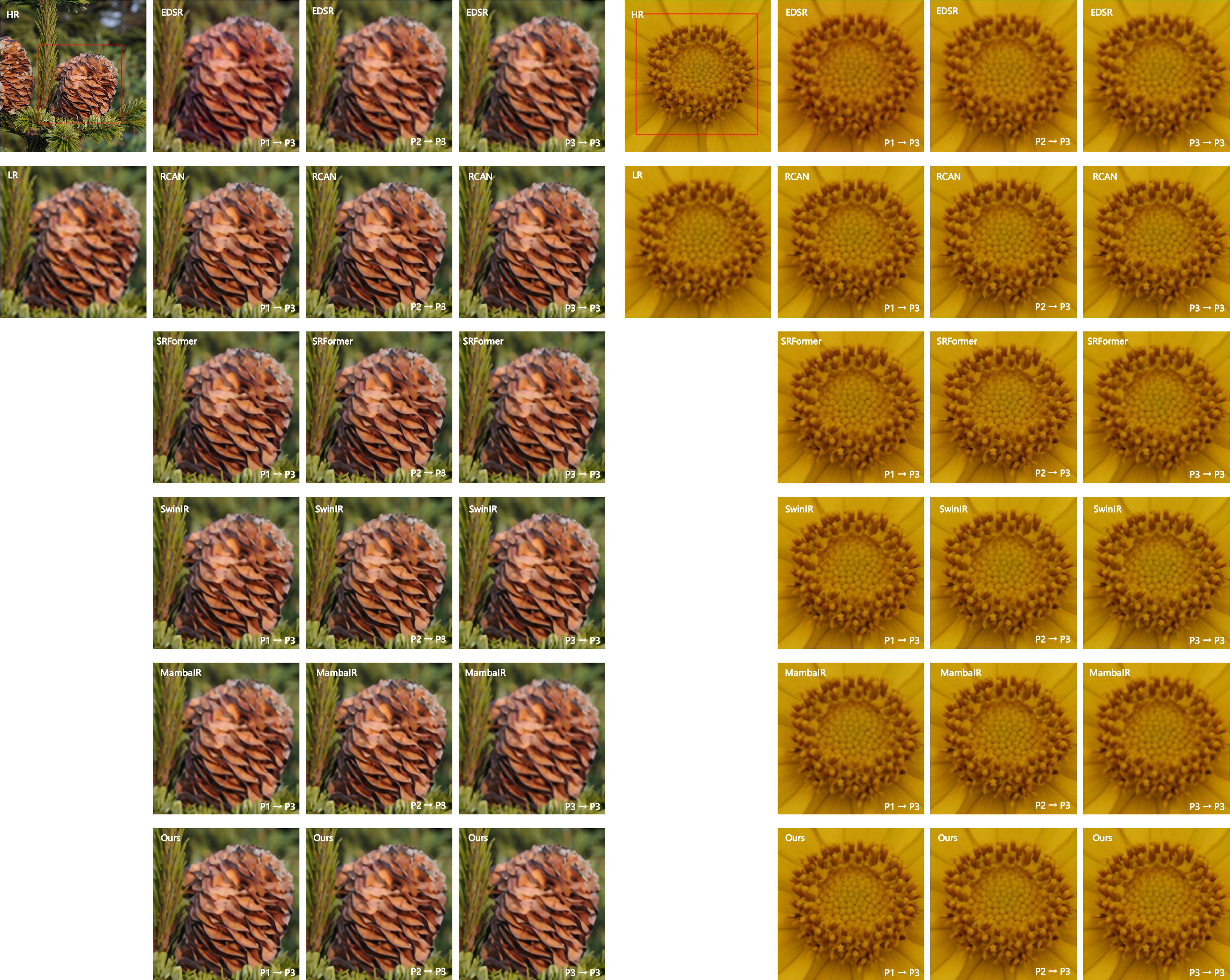}
        }
    \caption{Visual comparisons on Plant benchmarks for the SR task.} 
    \label{fig_bicubic_plant}
    \end{figure*}

\textbf{Comparision on Plant Benchmarks.}
Tab.~\ref{tab_sr_plant} establishes a cross-dataset evaluation framework using three Plant benchmarks: PlantSR (P1), LSDIR-plant (P2), and our PlantDR (P3). This design assesses  in-domain test such as P3 to P3, and cross-domain generalization test including P3 to P1. 
Quantitative evaluations show that our method achieves SOTA perceptual quality with only one-second the model size of SRFormer. 
This is evidenced by competitive or best-in-class LPIPS scores across all test cases in Table~\ref{tab_sr_plant}.
Meanwhile, our method maintains strong PSNR and SSIM performance, demonstrating a balanced trade-off between perceptual quality and distortion-based metrics.
Moreover, results show that our PlantDR dataset benefits not only our model but also improves baseline performance in both in-domain and cross-domain settings. Models trained on PlantDR achieve consistently better LPIPS and PSNR scores, indicating that its diverse, well-structured content enhances generalization and robustness.
Qualitative results are presented in Fig.~\ref{fig_bicubic_plant}. From the visual comparisons, our model produces similar perceptual quality to other methods despite its significantly smaller model size. Moreover, models trained on our proposed dataset exhibit noticeably improved visual performance, demonstrating the effectiveness of the dataset in enhancing perceptual quality.

\subsection{Performance of Image Denosing}
The five State-of-the-Art methods compared are DRUNet~\cite{DRUNet}, Xformer~\cite{Xformer}, RNINet~\cite{RNINet}, SwinIR~\cite{SwinIR}, MambaIR~\cite{MambaIR}.
Among them, DRUNet, Xformer, and RNINet are representative SOTA methods in the denoising field employing CNN and Transformer architectures, while SwinIR and MambaIR are general-purpose methods for Image Restoration utilizing Transformer and Mamba architectures, respectively.
For a fair comparison, We trained all the models from scratch using DF2K with additive white Gaussian noise~($\sigma \in [0,50]$). All other hyper-parameters were kept identical to their original configurations.
All experiments are conducted at a noise level of $\sigma = 25$.
We first evaluate on common benchmarks, and then compare on plant datasets.

\renewcommand{\arraystretch}{1.2} %
\begin{table}[h]

\centering

\small
\setlength{\tabcolsep}{2mm}
\resizebox{\linewidth}{!}{

\begin{tabular} {c|c|c|cccccc}
    \hline
    Datasets   & sigma &  Metrics &  Set5 & Set14 & B100 & Urban100 & Mamga109 & DIV2K-VAL \\  
\hline
\multirowcell{3}{\textbf{DRUNet}} & 25  &PSNR$\uparrow$ &33.74 &33.01 &32.51 &33.45 &33.56 &34.35	 \\
& 25   &SSIM$\uparrow$ &\textbf{0.9191} &\textbf{0.9042} &0.8972 &\textbf{0.9383} &0.9519 	&0.9216 \\
32.64M & 25     &LPIPS$\downarrow$  &0.0988 &0.1086 &0.1149 &0.0633 &0.0619 	&0.1394 \\

\hline
\multirowcell{3}{\textbf{Xformer}}& 25    &PSNR$\uparrow$ &\textbf{33.80} &\textbf{33.10} &32.60&\textbf{33.72 } &\textbf{33.61} 	&\textbf{34.47} \\
& 25 &SSIM$\uparrow$ &0.9092 &0.9018 &\textbf{0.9396} &0.8993 &0.9434 	&\textbf{0.9227} \\
25.23M & 25 &LPIPS$\downarrow$ &\textbf{0.0574} &0.0953 &\textbf{0.0479} &0.0958 &\textbf{0.0482} 	&\textbf{0.1181} \\

\hline
\multirowcell{3}{\textbf{RNINet}}& 25    &PSNR$\uparrow$ &33.52 &32.76 &32.37 &32.76 &33.16 	&34.14 \\
& 25 &SSIM$\uparrow$ &0.9175 &0.9010 &0.8969 &0.9314 &0.9433 &0.9200 \\
11.56M & 25 &LPIPS$\downarrow$  &0.0827 &0.1064 &0.1037 &0.0625 &0.05981 	&0.1280 \\
 
\hline
\multirowcell{3}{\textbf{SwinIR}}& 25    &PSNR$\uparrow$ &33.61 &33.02 &\textbf{33.33} &32.49 &33.48 	&34.29 \\
& 25 &SSIM$\uparrow$  &0.9149 &0.9039 &0.9346 &0.8970 &0.9435 	&0.9186 \\
13.29M & 25 &LPIPS$\downarrow$ &0.0744 &0.0989 &0.0603 &0.1048 &0.0603 	&0.1319 \\

\hline
\multirowcell{3}{\textbf{MambaIR}}& 25    &PSNR$\uparrow$  &33.59 &32.82 &32.47 &33.24 & 32.35	&34.27 \\
& 25 &SSIM$\uparrow$ &0.9161 &0.8956 &0.8972 & 0.9334& 0.9377	&0.9199 \\
20.57M & 25 &LPIPS$\downarrow$ &0.0592 &0.0998 &0.0999 & \textbf{0.0561}& 0.0546	& 0.1217\\

\hline
\multirowcell{3}{\textbf{Ours}}& 25    &PSNR$\uparrow$  &33.30 &32.50 &32.16 &32.40 &33.07 	& 33.84\\
& 25 &SSIM$\uparrow$ &0.9117 &0.8936 &0.8906 &0.9245 &\textbf{0.9441} 	&0.9139 \\
3.52M & 25 &LPIPS$\downarrow$ &0.0901 &0.1185 &0.1223 &0.0696 & 0.0599	&0.1420 \\

\hline

\end{tabular}{}
}
    \caption{ Quantitative comparison with state-of-the-art methods on common benchmarks  for the image denoising task.}
\label{tab_denoise_com}
\end{table}
\begin{figure*}[h]
    \centering
        \resizebox{\linewidth}{!}{
        \includegraphics[width=1\textwidth]{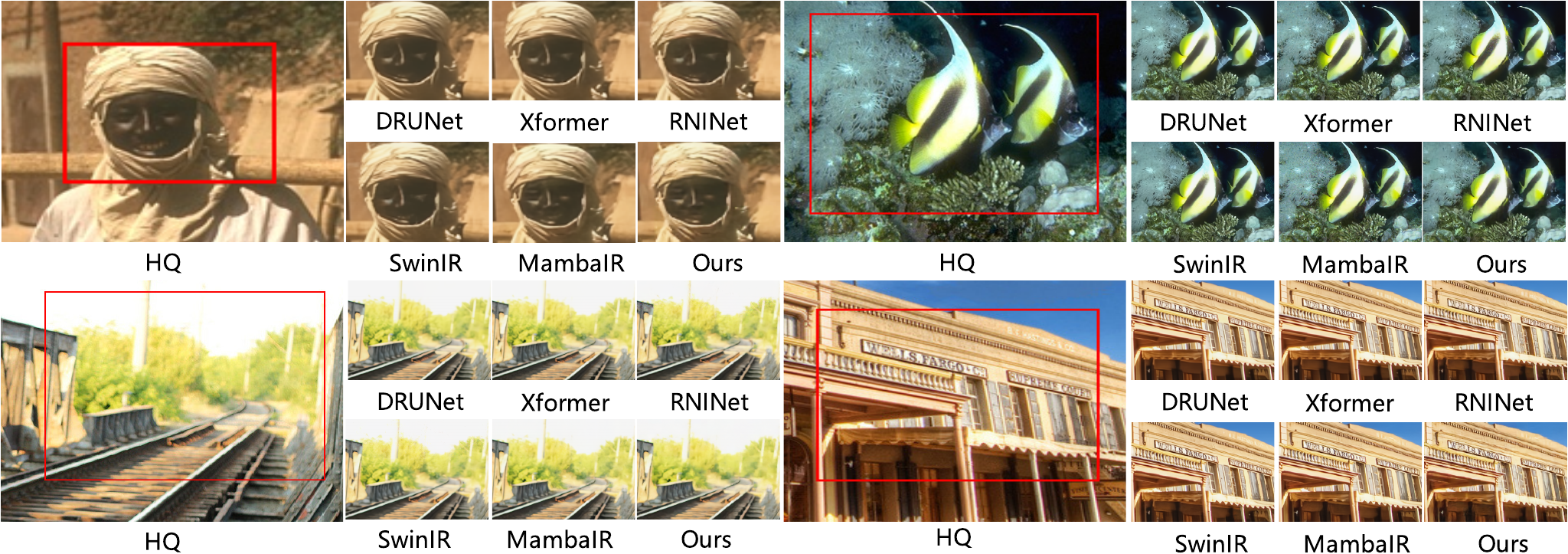}
        }
    \caption{Visual comparisons on common benchmarks for the image denoising task.} 
    \label{fig_denoise}
    \end{figure*}

\textbf{Comparision on Common Benchmarks.}
The quantitative and qualitative results are presented in Tab.~\ref{tab_denoise_com} and Fig.~\ref{fig_denoise}, respectively.
As listed in Table~\ref{tab_denoise_com}, our model achieves competitive denoising performance while using significantly fewer parameters compared to other SOTA methods. Specifically, our model has only 3.52M parameters, which is substantially smaller than DRUNet (32.64M), Xformer (25.23M), RNINet (11.56M), SwinIR (13.29M), and MambaIR (20.57M). Despite this, our method delivers comparable results across most benchmarks. For example, it achieves a PSNR of 33.30 on Set5 and 33.84 on DIV2K-VAL, which are close to those of larger models. Notably, our model attains the highest SSIM of 0.9441 on Manga109 and competitive LPIPS scores on several datasets. These results demonstrate that our method strikes a good balance between model size and performance, making it an efficient and effective solution for image denoising.
From the qualitative results, our model achieves a better balance between preserving fine textures and effectively removing noise. Compared to other methods, it retains more structural details and avoids over-smoothing, while still suppressing noise artifacts. 

\renewcommand{\arraystretch}{1.2} %
\begin{table}[!ht]

\centering

\small
\setlength{\tabcolsep}{2mm}
\resizebox{\linewidth}{!}{

\begin{tabular} {c|c|c|ccc|ccc}
    \hline
    Train $\rightarrow$ Test   & Sigma & Metrics &  P1 $\rightarrow$ P1 & P2 $\rightarrow$ P1 & P3 $\rightarrow$ P1 & P1 $\rightarrow$ P3 &  P2 $\rightarrow$ P3 & P3 $\rightarrow$ P3\\   
\hline
\multirowcell{3}{\textbf{DRUNet}} & 25  &PSNR$\uparrow$ &35.28 &35.11 &\textbf{35.35}  &37.44&37.32 &\textbf{37.52}	 \\
& 25   &SSIM$\uparrow$ &0.9361 &0.9349 &\textbf{0.9374}  	&0.9475& 0.9468&\textbf{0.9491}\\
32.64M v& 25     &LPIPS$\downarrow$  &0.1294 &\textbf{0.1188} &0.1275 	&0.1175&\textbf{0.1052} &0.1135 \\

\hline
\multirowcell{3}{\textbf{Xformer}}& 25    &PSNR$\uparrow$  &35.56 &32.23 &\textbf{ 35.58} 	&38.03&37.83 &\textbf{38.06} \\
& 25 &SSIM$\uparrow$  &0.9394 &0.9257 & \textbf{0.9403} 	&0.9541&0.9508 &\textbf{0.9544} \\
25.23M & 25 &LPIPS$\downarrow$  &\textbf{0.0990} &0.1144 &  0.1006	&0.0946&\textbf{0.0940} &0.0989 \\

\hline
\multirowcell{3}{\textbf{RNINet}}& 25    &PSNR$\uparrow$ &35.53 &35.36 &\textbf{35.54}  &\textbf{37.85} &37.66 &\textbf{37.85} \\
& 25 &SSIM$\uparrow$  &\textbf{0.9413} &0.9379 &0.9409 &\textbf{0.9528} &0.9504 &0.9523 \\
11.56M & 25 &LPIPS$\downarrow$   &0.1098 &0.1112 &\textbf{0.1087} &0.1066 &\textbf{0.1043} &0.1050 \\
 
\hline
\multirowcell{3}{\textbf{SwinIR}}& 25    &PSNR$\uparrow$ &35.51 &35.20 &\textbf{35.52} &37.72&37.30 &\textbf{37.82} \\
& 25 &SSIM$\uparrow$   &0.9372 &0.9193 &\textbf{0.9379}  	&0.9487&0.9385 &\textbf{0.9508} \\
13.29M & 25 &LPIPS$\downarrow$  &0.1368 &0.1837 &\textbf{0.1328} 	&0.1112&0.1282 &\textbf{0.1061} \\

\hline
\multirowcell{3}{\textbf{MambaIR}}& 25    &PSNR$\uparrow$  &35.40 &34.27 &\textbf{35.47} &37.41 &36.97 	&\textbf{37.43} \\
& 25 &SSIM$\uparrow$ &0.9327 &0.8838  &\textbf{0.9373} 	&0.9454 &0.9358 &\textbf{0.9461}\\
20.57M & 25 &LPIPS$\downarrow$ &0.1188 &0.2089  &\textbf{0.1063} 	&0.1148&0.1185 &\textbf{0.1128}\\

\hline
\multirowcell{3}{\textbf{Ours}}& 25    &PSNR$\uparrow$  &35.26 &35.09 &\textbf{35.30 }	&36.88&36.69 &\textbf{36.89} \\
& 25 &SSIM$\uparrow$ &\textbf{0.9365} &0.9341 &0.9363  	&0.9397&0.9376 &\textbf{0.9405} \\
3.52M & 25 &LPIPS$\downarrow$ &0.1158 &\textbf{0.1144} & 0.1200	&0.1236&\textbf{0.1184} &0.1245 \\

\hline

\end{tabular}{}
}
    \caption{ Quantitative comparison with state-of-the-art methods on plant benchmarks for the image denoising task.  P1 represents PlantSR~\cite{PlantSR}, P2 represent LSDIR-plant~\cite{LSDIR} and P3 is our proposed PlantDR.}
\label{tab_denoise_plant}
\end{table}

\textbf{Comparision on Plant Benchmarks.}
Based on the results presented in Table~\ref{tab_denoise_plant}, our model demonstrates strong denoising performance with the fewest parameters among all compared methods. Specifically, our method contains only 3.52M parameters, which is significantly smaller than DRUNet (32.64M), Xformer (25.23M), RNINet (11.56M), SwinIR (13.29M), and MambaIR (20.57M). Despite the lightweight design, our model achieves competitive results in both in-domain (e.g., P3$\rightarrow$P3) and cross-domain (e.g., P3$\rightarrow$P1) settings. 
For instance, our method achieves a PSNR of 36.89 and SSIM of 0.9405 on P3$\rightarrow$P3, which are very close to those of larger models such as RNINet and Xformer.
When evaluated out-of-domain (P3$\rightarrow$P1), our method still reaches 35.30 dB PSNR and 0.9363 SSIM, outperforming the same-sized SwinIR and MambaIR under the P1$\rightarrow$P1 condition.
Moreover, our newly proposed dataset, PlantDR (P3), proves to be beneficial for improving model generalization. 
For instance, DRUNet improves from 35.28 dB (P1$\rightarrow$P1) to 35.35 dB (P3$\rightarrow$P1), and RNINet rises from 35.53 dB (P1$\rightarrow$P1) to 35.54 dB (P3$\rightarrow$P1).
These gains underscore the superior diversity and representativeness of PlantDR, validating both the effectiveness of our lightweight architecture and the generalization benefits conferred by the dataset.

\renewcommand{\arraystretch}{1.2} %
\begin{table}[h]

\centering

\small
\setlength{\tabcolsep}{2mm}

\begin{tabular} {c|ccc}
    \hline
    Methods   &Params& FLOPS & MACs \\  
\hline
\textbf{EDSR} &1.52M &16.33G &8.12G \\
\hline

\textbf{RCAN}&15.59M  &130.62G &65.2G\\
\hline

\textbf{SRFormer}&10.54M  &112.91G &56.09G\\
\hline

\textbf{SwinIR}&13.29M  & 118.48G& 59.02G\\
\hline

\textbf{DRUNet} &32.64M  &8.67G &17.35G\\
\hline

\textbf{Xformer}&25.23M  &20.55G &10.25G\\
\hline

\textbf{RNINet}&11.56M  &2.48G &4.97G\\
\hline

\textbf{MambaIR}&20.57M  &164.06G &69.75G\\
\hline

\textbf{Ours}&3.52M  & 35.95G&15.17G\\
\hline

\end{tabular}{}
    \caption{ Comparison of model size and computational cost.}
\label{tab_comp}
\end{table}

\subsection{Efficiency Comparison}
Based on the results in Table~\ref{tab_comp}, our model demonstrates a favorable balance between model size and computational efficiency. Specifically, it has only 3.52M parameters, which is significantly smaller than most competing methods such as DRUNet (32.64M), RCAN (15.59M), and SwinIR (13.29M). In terms of FLOPs and MACs, our method also shows moderate computational cost at 35.95G FLOPs and 15.17G MACs, offering a good trade-off between efficiency and performance.
Compared to lightweight models such as RNINet (2.48G FLOPs) and DRUNet (8.67G FLOPs), our method has slightly higher computational cost but offers improved representational capacity and performance, as listed in \Cref{tab_sr_com,tab_sr_plant,tab_denoise_com,tab_denoise_plant}. On the other hand, our model is significantly more efficient than transformer-based models such as SRFormer (112.91G FLOPs), SwinIR (118.48G FLOPs), and Mamba-based method MambaIR (164.06G FLOPs), while maintaining comparable or better reconstruction quality.
These results suggest that our model is not only compact in terms of parameter count, but also computationally efficient, making it well-suited for real-world applications that require both high performance and low resource consumption.
\setlength{\tabcolsep}{1.2pt} %
\renewcommand{\arraystretch}{1} %
\begin{table}[t]
\centering

\small

      \centering
      \small
      \setlength{\tabcolsep}{2mm}
      \begin{tabular}{c|c|c|c}
        \hline
        Setting & PSNR$\uparrow$ & SSIM$\uparrow$ & LPIPS$\downarrow$ \\
        \hline
         w/o Depth & 30.24 & 0.8324 & 0.2800 \\
        \hline
        w Depth & \textbf{30.28} & \textbf{0.8335} & \textbf{0.2787} \\
        \hline
      \end{tabular}
\caption{Ablation study of dual branch settings.}
      \label{tab_abDual}
\end{table}
\begin{figure*}[t]
    \centering
        \resizebox{\linewidth}{!}{
        \includegraphics[width=1\textwidth]{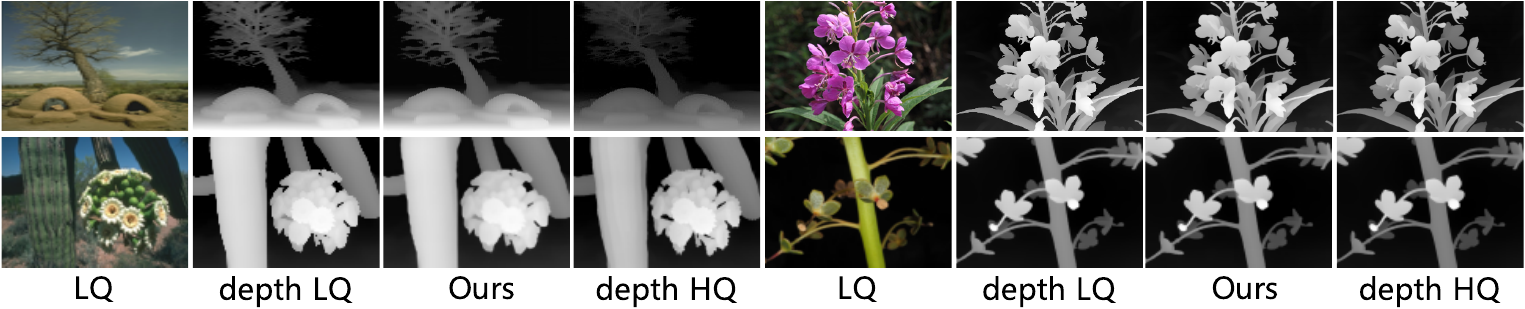}
        }
    \caption{Visualization of reconstructed depth maps.} 
    \label{fig_depth}
    \end{figure*}

\section{Ablation Study}
\textbf{Ablation on Depth-Free Architecture.} 
We present ablation results on SR without the depth branch in Tab.~\ref{tab_abDual}. Disabling depth branch yields consistent drops across all three metrics. Although the model still converges to a competitive level of overall performance, the absence of depth awareness prevents it from concentrating on foreground regions that are perceptually salient to human observers.

\textbf{Ablation on Depth-Free Architecture.} 
We present ablation results on SR without the depth branch in Tab.~\ref{tab_abDual}. Disabling depth branch yields consistent drops across all three metrics. Although the model still converges to a competitive level of overall performance, the absence of depth awareness prevents it from concentrating on foreground regions that are perceptually salient to human observers.
The effectiveness of our depth reconstruction is demonstrated in Fig.~\ref{fig_depth}, where it can be seen that our model successfully recovers low-quality depth maps with high fidelity. These reconstructed depth maps serve as informative structural priors, effectively guiding and enhancing the performance of the image reconstruction branch.

\begin{figure*}[t]
    \centering
        \resizebox{\linewidth}{!}{
        \includegraphics[width=1\textwidth]{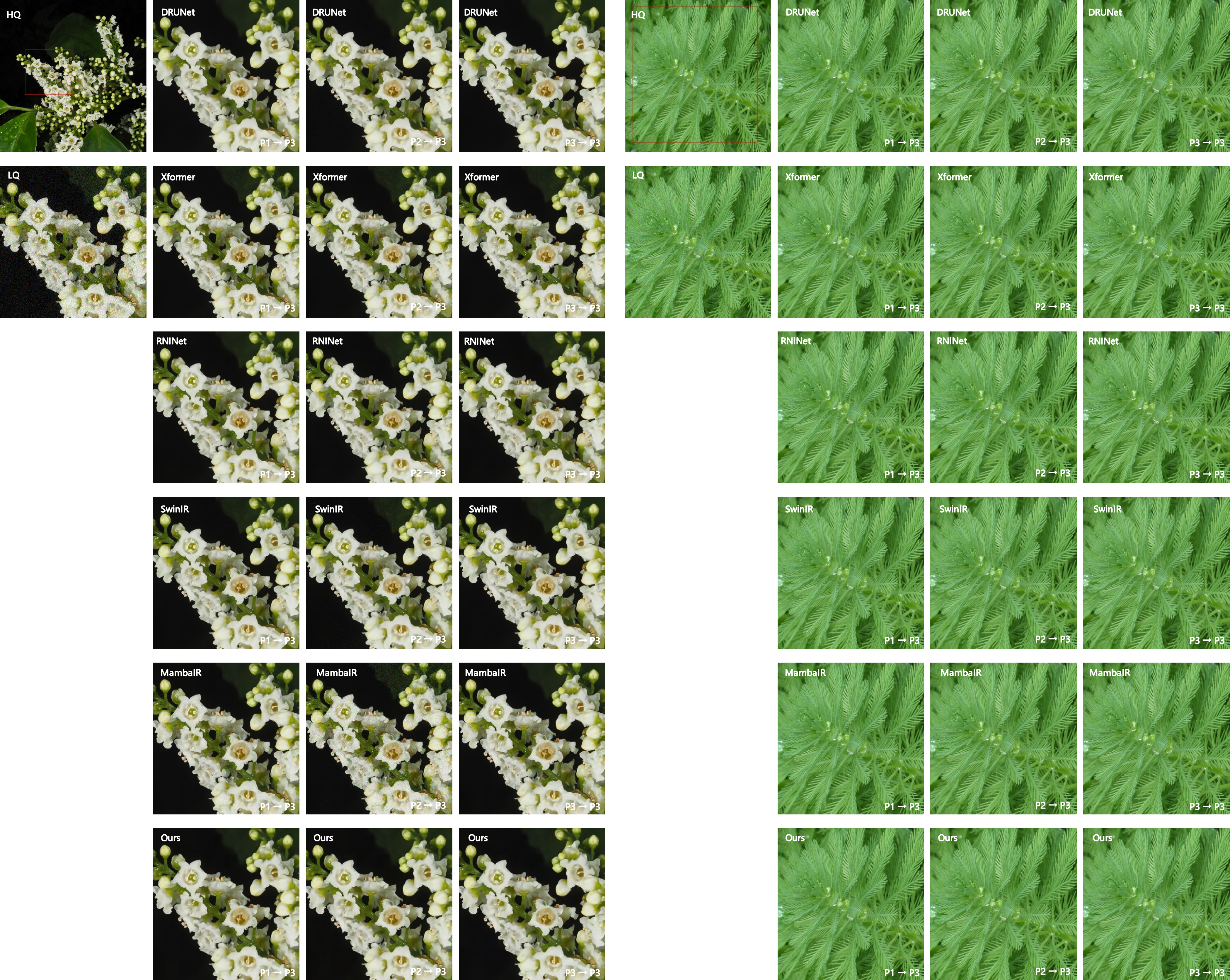}
        }
    \caption{Visual comparisons on Plant benchmarks for the image denoising task.} 
    \label{fig_denoise_plant}
    \end{figure*}

\section{Conclusion}
We propose a novel depth-guided network for image restoration, together with a novel large-sacel high-resolution dataset.
The network consists of two interactive branches: a depth estimation branch that provides structural guidance, and an image restoration branch that performs the core restoration task. The image restoration branch exploits intra-object similarity through progressive window-based self-attention and captures inter-object similarity via sparse non-local attention. Through joint training, depth features contribute to improved restoration quality, while the enhanced visual features from the restoration branch in turn help refine depth estimation.
We also introduce a new dataset for training and evaluation, consisting of 9,205 high-resolution images from 403 plant species, with diverse depth and texture variations.
Extensive experiments show that our method achieves state-of-the-art performance on several standard benchmarks and generalizes well to previously unseen plant images, demonstrating both its effectiveness and robustness.

{
\small
\bibliography{aaai2026}

\begin{thebibliography}{40}
\providecommand{\natexlab}[1]{#1}

\bibitem[{Agustsson and Timofte(2017)}]{DIV2K}
Agustsson, E.; and Timofte, R. 2017.
\newblock {NTIRE} 2017 Challenge on Single Image Super-Resolution: Dataset and
  Study.
\newblock In \emph{{IEEE} Conference on Computer Vision and Pattern Recognition
  Workshops}, 1122--1131.

\bibitem[{Bevilacqua et~al.(2012)Bevilacqua, Roumy, Guillemot, and
  Alberi-Morel}]{Set5}
Bevilacqua, M.; Roumy, A.; Guillemot, C.; and Alberi-Morel, M.-L. 2012.
\newblock Low-Complexity Single-Image Super-Resolution based on Nonnegative
  Neighbor Embedding.
\newblock In \emph{British Machine Vision Conference}, 1--10.

\bibitem[{Chen et~al.(2023)Chen, Liang, Zhang, Liu, Zeng, and Zhang}]{HGGT}
Chen, D.; Liang, J.; Zhang, X.; Liu, M.; Zeng, H.; and Zhang, L. 2023.
\newblock Human Guided Ground-Truth Generation for Realistic Image
  Super-Resolution.
\newblock In \emph{{IEEE} Conference on Computer Vision and Pattern
  Recognition}, 14082--14091.

\bibitem[{Dai et~al.(2019)Dai, Cai, Zhang, Xia, and Zhang}]{SAN}
Dai, T.; Cai, J.; Zhang, Y.; Xia, S.; and Zhang, L. 2019.
\newblock Second-Order Attention Network for Single Image Super-Resolution.
\newblock In \emph{{IEEE} Conference on Computer Vision and Pattern
  Recognition}, 11065--11074.

\bibitem[{Dosovitskiy et~al.(2021)Dosovitskiy, Beyer, Kolesnikov, Weissenborn,
  Zhai, Unterthiner, Dehghani, Minderer, Heigold, Gelly, Uszkoreit, and
  Houlsby}]{ViT}
Dosovitskiy, A.; Beyer, L.; Kolesnikov, A.; Weissenborn, D.; Zhai, X.;
  Unterthiner, T.; Dehghani, M.; Minderer, M.; Heigold, G.; Gelly, S.;
  Uszkoreit, J.; and Houlsby, N. 2021.
\newblock An Image is Worth 16x16 Words: Transformers for Image Recognition at
  Scale.
\newblock In \emph{International Conference on Learning Representations}.

\bibitem[{Gu and Dao(2024)}]{Mamba}
Gu, A.; and Dao, T. 2024.
\newblock Mamba: Linear-Time Sequence Modeling with Selective State Spaces.
\newblock In \emph{Conference on Language Modeling}.

\bibitem[{Gu et~al.(2019)Gu, Lugmayr, Danelljan, Fritsche, Lamour, and
  Timofte}]{DIV8K}
Gu, S.; Lugmayr, A.; Danelljan, M.; Fritsche, M.; Lamour, J.; and Timofte, R.
  2019.
\newblock {DIV8K:} DIVerse 8K Resolution Image Dataset.
\newblock In \emph{International Conference on Computer Vision Workshops},
  3512--3516.

\bibitem[{Guo et~al.(2024)Guo, Li, Dai, Ouyang, Ren, and Xia}]{MambaIR}
Guo, H.; Li, J.; Dai, T.; Ouyang, Z.; Ren, X.; and Xia, S. 2024.
\newblock {MambaIR}: {A} Simple Baseline for Image Restoration with State-Space
  Model.
\newblock In \emph{European conference on computer vision}, 222--241.

\bibitem[{Huang, Singh, and Ahuja(2015)}]{Urban100}
Huang, J.-B.; Singh, A.; and Ahuja, N. 2015.
\newblock Single image super-resolution from transformed self-exemplars.
\newblock In \emph{{IEEE} Conference on Computer Vision and Pattern
  Recognition}, 5197--5206.

\bibitem[{Jiang et~al.(2024)Jiang, Yu, Zhong, and Shao}]{PlantSR}
Jiang, T.; Yu, Q.; Zhong, Y.; and Shao, M. 2024.
\newblock PlantSR: Super-Resolution Improves Object Detection in Plant Images.
\newblock \emph{Journal of Imaging}, 10(6): 137.

\bibitem[{Kim, Sim, and Kim(2021)}]{KOALAnet}
Kim, S.~Y.; Sim, H.; and Kim, M. 2021.
\newblock KOALAnet: Blind Super-Resolution Using Kernel-Oriented Adaptive Local
  Adjustment.
\newblock In \emph{{IEEE} Conference on Computer Vision and Pattern
  Recognition}, 10611--10620.

\bibitem[{Kingma and Ba(2015)}]{ADAM}
Kingma, D.~P.; and Ba, J. 2015.
\newblock Adam: {A} Method for Stochastic Optimization.
\newblock In \emph{International Conference on Learning Represent}.

\bibitem[{Li et~al.(2023)Li, Zhang, Liang, Cao, Liu, Gong, Zhang, Tang, Liu,
  Demandolx, Ranjan, Timofte, and Gool}]{LSDIR}
Li, Y.; Zhang, K.; Liang, J.; Cao, J.; Liu, C.; Gong, R.; Zhang, Y.; Tang, H.;
  Liu, Y.; Demandolx, D.; Ranjan, R.; Timofte, R.; and Gool, L.~V. 2023.
\newblock {LSDIR:} {A} Large Scale Dataset for Image Restoration.
\newblock In \emph{IEEE Conference on Computer Vision and Pattern Recognition
  Workshops}, 1775--1787.

\bibitem[{Liang et~al.(2021)Liang, Cao, Sun, Zhang, Van~Gool, and
  Timofte}]{SwinIR}
Liang, J.; Cao, J.; Sun, G.; Zhang, K.; Van~Gool, L.; and Timofte, R. 2021.
\newblock {SwinIR}: Image restoration using swin transformer.
\newblock In \emph{IEEE International Conference on Computer Vision},
  1833--1844.

\bibitem[{Lim et~al.(2017)Lim, Son, Kim, Nah, and Mu~Lee}]{MDSR}
Lim, B.; Son, S.; Kim, H.; Nah, S.; and Mu~Lee, K. 2017.
\newblock Enhanced deep residual networks for single image super-resolution.
\newblock In \emph{{IEEE} Conference on Computer Vision and Pattern Recognition
  Workshops}, 136--144.

\bibitem[{Liu et~al.(2018)Liu, Jaw, Huang, and Hwang}]{DesnowNet}
Liu, Y.; Jaw, D.; Huang, S.; and Hwang, J. 2018.
\newblock DesnowNet: Context-Aware Deep Network for Snow Removal.
\newblock \emph{IEEE Transactions on Image Processing}, 27(6): 3064--3073.

\bibitem[{Lyu et~al.(2025)Lyu, Cha, Jiang, Chowdury, Hou, Harake, Kondepudi,
  Freudiger, Lee, and Hollon}]{DBLP:conf/aaai/LyuC0CHHKFLH25}
Lyu, Y.; Cha, S.~J.; Jiang, C.; Chowdury, A.~Z.; Hou, X.; Harake, E.~S.;
  Kondepudi, A.; Freudiger, C.~W.; Lee, H.; and Hollon, T.~C. 2025.
\newblock Step-Calibrated Diffusion for Biomedical Optical Image Restoration.
\newblock In \emph{Proceedings of the AAAI Conference on Artificial
  Intelligence}, 5946--5954.

\bibitem[{Martin et~al.(2001)Martin, Fowlkes, Tal, and Malik}]{B100}
Martin, D.; Fowlkes, C.; Tal, D.; and Malik, J. 2001.
\newblock A database of human segmented natural images and its application to
  evaluating segmentation algorithms and measuring ecological statistics.
\newblock In \emph{IEEE International Conference on Computer Vision}, 416--425.

\bibitem[{Mei, Fan, and Zhou(2021)}]{NLSN}
Mei, Y.; Fan, Y.; and Zhou, Y. 2021.
\newblock Image Super-Resolution With Non-Local Sparse Attention.
\newblock In \emph{{IEEE} Conference on Computer Vision and Pattern
  Recognition}, 3517--3526.

\bibitem[{Monga and Evans(2006)}]{zauner2010phash}
Monga, V.; and Evans, B.~L. 2006.
\newblock Perceptual Image Hashing Via Feature Points: Performance Evaluation
  and Tradeoffs.
\newblock \emph{{IEEE} Transactions on Image Processing}, 15(11): 3452--3465.

\bibitem[{Wang et~al.(2025)Wang, Yan, Niu, Gong, Guo, and Wei}]{DEMoreNet}
Wang, Y.; Yan, X.; Niu, Y.; Gong, L.; Guo, Y.; and Wei, M. 2025.
\newblock Rethinking mixture of rain removal via depth-guided adversarial
  learning.
\newblock \emph{Neural Networks}, 191: 107739.

\bibitem[{Wang et~al.(2004)Wang, Bovik, Sheikh, and Simoncelli}]{SSIM}
Wang, Z.; Bovik, A.~C.; Sheikh, H.~R.; and Simoncelli, E.~P. 2004.
\newblock Image quality assessment: from error visibility to structural
  similarity.
\newblock \emph{IEEE Transactions on Image Processing}, 13(4): 600--612.

\bibitem[{Yang et~al.(2024{\natexlab{a}})Yang, Kang, Huang, Xu, Feng, and
  Zhao}]{DepthAnything}
Yang, L.; Kang, B.; Huang, Z.; Xu, X.; Feng, J.; and Zhao, H.
  2024{\natexlab{a}}.
\newblock Depth Anything: Unleashing the Power of Large-Scale Unlabeled Data.
\newblock In \emph{{IEEE/CVF} Conference on Computer Vision and Pattern
  Recognition}, 10371--10381.

\bibitem[{Yang et~al.(2024{\natexlab{b}})Yang, Kang, Huang, Zhao, Xu, Feng, and
  Zhao}]{DepthAnythingV2}
Yang, L.; Kang, B.; Huang, Z.; Zhao, Z.; Xu, X.; Feng, J.; and Zhao, H.
  2024{\natexlab{b}}.
\newblock Depth Anything {V2}.
\newblock In \emph{Advances in Neural Information Processing Systems}.

\bibitem[{Yang et~al.(2024{\natexlab{c}})Yang, Wang, Guo, and Tao}]{D4}
Yang, Y.; Wang, C.; Guo, X.; and Tao, D. 2024{\natexlab{c}}.
\newblock Robust Unpaired Image Dehazing via Density and Depth Decomposition.
\newblock \emph{International Journal of Computer Vision}, 132(5): 1557--1577.

\bibitem[{Yin et~al.(2025)Yin, Wang, Lin, Ran, and Zheng}]{RNINet}
Yin, Z.; Wang, H.; Lin, G.; Ran, W.; and Zheng, Y. 2025.
\newblock Random Is All You Need: Random Noise Injection on Feature Statistics
  for Generalizable Deep Image Denoising.
\newblock In \emph{International Conference on Learning Representations}.

\bibitem[{Yusuke et~al.(2016)Yusuke, Kota, Yuji, Azuma, Toru, Toshihiko, and
  Kiyoharu}]{Manga109}
Yusuke, M.; Kota, I.; Yuji, A.; Azuma, F.; Toru, O.; Toshihiko, Y.; and
  Kiyoharu, A. 2016.
\newblock Sketch-based manga retrieval using manga109 dataset.
\newblock \emph{Multimedia Tools and Applications}, 68(8): 21811--21838.

\bibitem[{Zeyde, Elad, and Protter(2010)}]{Set14}
Zeyde, R.; Elad, M.; and Protter, M. 2010.
\newblock On Single Image Scale-Up Using Sparse-Representations.
\newblock In \emph{Curves and Surfaces}, 711--730.

\bibitem[{Zhang et~al.(2024{\natexlab{a}})Zhang, Zhang, Gu, Dong, Kong, and
  Yang}]{Xformer}
Zhang, J.; Zhang, Y.; Gu, J.; Dong, J.; Kong, L.; and Yang, X.
  2024{\natexlab{a}}.
\newblock Xformer: Hybrid X-Shaped Transformer for Image Denoising.
\newblock In \emph{International Conference on Learning Representations}.

\bibitem[{Zhang et~al.(2022{\natexlab{a}})Zhang, Li, Zuo, Zhang, Gool, and
  Timofte}]{DRUNet}
Zhang, K.; Li, Y.; Zuo, W.; Zhang, L.; Gool, L.~V.; and Timofte, R.
  2022{\natexlab{a}}.
\newblock Plug-and-Play Image Restoration With Deep Denoiser Prior.
\newblock \emph{IEEE Transactions on Pattern Analysis and Machine
  Intelligence}, 44(10): 6360--6376.

\bibitem[{Zhang et~al.(2024{\natexlab{b}})Zhang, Gong, Jiang, Shi, Yang, Bao,
  and Zhi}]{ZHANG2024107262}
Zhang, P.; Gong, J.; Jiang, S.; Shi, T.; Yang, J.; Bao, G.; and Zhi, X.
  2024{\natexlab{b}}.
\newblock A method for remote sensing image restoration based on the system
  degradation model.
\newblock \emph{Results in Physics}, 56: 107262.

\bibitem[{Zhang et~al.(2018{\natexlab{a}})Zhang, Isola, Efros, Shechtman, and
  Wang}]{LPIPS}
Zhang, R.; Isola, P.; Efros, A.~A.; Shechtman, E.; and Wang, O.
  2018{\natexlab{a}}.
\newblock The unreasonable effectiveness of deep features as a perceptual
  metric.
\newblock In \emph{{IEEE} Conference on Computer Vision and Pattern
  Recognition}, 586--595.

\bibitem[{Zhang et~al.(2024{\natexlab{c}})Zhang, Yang, Zhu, Zhou, Wang, and
  Yin}]{zsx3}
Zhang, S.-X.; Yang, C.; Zhu, X.; Zhou, H.; Wang, H.; and Yin, X.-C.
  2024{\natexlab{c}}.
\newblock Inverse-like antagonistic scene text spotting via reading-order
  estimation and dynamic sampling.
\newblock \emph{IEEE Transactions on Image Processing}, 33: 825--839.

\bibitem[{Zhang et~al.(2022{\natexlab{b}})Zhang, Zhu, Chen, Hou, and
  Yin}]{zsx1}
Zhang, S.-X.; Zhu, X.; Chen, L.; Hou, J.-B.; and Yin, X.-C. 2022{\natexlab{b}}.
\newblock Arbitrary shape text detection via segmentation with probability
  maps.
\newblock \emph{IEEE transactions on pattern analysis and machine
  intelligence}, 45(3): 2736--2750.

\bibitem[{Zhang et~al.(2022{\natexlab{c}})Zhang, Zhu, Hou, Yang, and
  Yin}]{zsx2}
Zhang, S.-X.; Zhu, X.; Hou, J.-B.; Yang, C.; and Yin, X.-C. 2022{\natexlab{c}}.
\newblock Kernel proposal network for arbitrary shape text detection.
\newblock \emph{IEEE transactions on neural networks and learning systems},
  34(11): 8731--8742.

\bibitem[{Zhang, Zhang, and Yu(2024)}]{HiTSR}
Zhang, X.; Zhang, Y.; and Yu, F. 2024.
\newblock HiT-SR: Hierarchical Transformer for Efficient Image
  Super-Resolution.
\newblock In \emph{European conference on computer vision}, volume 15098,
  483--500.

\bibitem[{Zhang et~al.(2018{\natexlab{b}})Zhang, Li, Li, Wang, Zhong, and
  Fu}]{RCAN}
Zhang, Y.; Li, K.; Li, K.; Wang, L.; Zhong, B.; and Fu, Y. 2018{\natexlab{b}}.
\newblock Image Super-Resolution Using Very Deep Residual Channel Attention
  Networks.
\newblock In \emph{European conference on computer vision}, 294--310.

\bibitem[{Zhou et~al.(2025)Zhou, Zhu, Qin, Xu, Cesar-Jr, and Yin}]{zhy2}
Zhou, H.; Zhu, X.; Qin, J.; Xu, Y.; Cesar-Jr, R.~M.; and Yin, X.-C. 2025.
\newblock Multi-Scale Texture Fusion for Reference-Based Image
  Super-Resolution: New Dataset and Solution.
\newblock \emph{International Journal of Computer Vision}, 1--22.

\bibitem[{Zhou et~al.(2023{\natexlab{a}})Zhou, Zhu, Zhu, Han, Zhang, Qin, and
  Yin}]{zhy1}
Zhou, H.; Zhu, X.; Zhu, J.; Han, Z.; Zhang, S.-X.; Qin, J.; and Yin, X.-C.
  2023{\natexlab{a}}.
\newblock Learning correction filter via degradation-adaptive regression for
  blind single image super-resolution.
\newblock In \emph{Proceedings of the IEEE/CVF International Conference on
  Computer Vision}, 12365--12375.

\bibitem[{Zhou et~al.(2023{\natexlab{b}})Zhou, Li, Guo, Bai, Cheng, and
  Hou}]{SRFormer}
Zhou, Y.; Li, Z.; Guo, C.; Bai, S.; Cheng, M.; and Hou, Q. 2023{\natexlab{b}}.
\newblock SRFormer: Permuted Self-Attention for Single Image Super-Resolution.
\newblock In \emph{International Conference on Computer Vision}, 12734--12745.

\end{thebibliography}
}

\end{document}